\documentclass[letterpaper]{article}
\usepackage{times} 
\usepackage{helvet} 
\usepackage{courier} 
\usepackage[hyphens]{url} 
\usepackage{graphicx} 
\usepackage{graphicx} 
\usepackage{natbib} 
\usepackage{caption} 
\usepackage{amssymb,amsmath,amsfonts, amsthm}
\usepackage{algorithm, algpseudocode}
\newtheorem{theorem}{Theorem}
\newtheorem{condition}{Condition}
\newtheorem{lemma}[theorem]{Lemma}
\title{Joint Scoring Rules:\\ Zero-Sum Competition Avoids Performative Prediction}
\author{Rubi Hudson\thanks{University of Toronto}}

\begin{document}
\maketitle
\begin{abstract}
In a decision-making scenario, a principal could use conditional predictions from an expert agent to inform their choice. However, this approach would introduce a fundamental conflict of interest. An agent optimizing for predictive accuracy is incentivized to manipulate their principal towards more predictable actions, which prevents that principal from being able to deterministically select their true preference. We demonstrate that this impossibility result can be overcome through the joint evaluation of multiple agents. When agents are made to engage in zero-sum competition, their incentive to influence the action taken is eliminated, and the principal can identify and take the action they most prefer. We further prove that this zero-sum setup is unique, efficiently implementable, and applicable under stochastic choice. Experiments in a toy environment demonstrate that training on a zero-sum objective significantly enhances both predictive accuracy and principal utility, and can eliminate previously learned manipulative behavior.
\end{abstract}

\section{Introduction}
Large Language Models (LLMs) excel at pattern recognition and predicting text. As capabilities improve, this predictive ability will be applied more frequently to real-world outcomes. This could include predicting human preferences, as is used in reinforcement learning from human feedback (RLHF) \citep{christiano2017reinforcement}, or predicting outcomes on which to give feedback in environments where only one action can be taken. Restricting highly capable AI systems to only make predictions, rather than autonomously pursue goals, is one proposal for AI alignment known as Oracle AI (\citealp{armstrong2012thinking}; \citealp{armstrong2013risks}).\\
A potential risk of using AI to make predictions is that the very act of making a prediction can affect the outcome, in a phenomenon known as performative prediction \citep{perdomo2020performative}. For example, predictions of inflation affect consumer behavior and thus inflation, predictions of crime affect police deployments and thus crime, and predictions of mortality affect healthcare usage and thus mortality. \\
When non-performative predictions are evaluated using a strictly proper scoring rule, the only optimal response is honesty \citep{gneiting2007strictly}. However, all such rules give higher expected scores when the underlying distribution is more extreme. In the performative case, using such a rule then incentives influencing the resulting distribution to be more extreme. \citep{oesterheld2023incentivizing}.\\
This introduces two risks. The first is that such manipulation could push the world towards becoming more predictable, with potentially negative effects on inflation, crime, healthcare, and other important outcomes. In the worst case, it could pose an existential risk — a dead human is much more predictable than a live one. \citep{hubinger2023conditiong}. The second risk is that optimal performative predictions are typically not stable upon reflection \citep{oesterheld2023incentivizing}, which drastically lowers their usefulness for decision making or training powerful AI systems \\
Still, one might ask how performative prediction could arise if we do not intentionally run a gradient running through a prediction's impact on outcomes? We identify four main pathways by which this could occur:
\begin{enumerate}
    \item Performative prediction could occur if a model implements search, either by design or as learned behavior \citep{hubinger2019risks}, and chooses between predictions based on expected score.
    \item Performative prediction could result from selection pressures other than gradient descent, such as population based training \citep{krueger2020hidden} or economic competition \citep{hendrycks2023natural}.
    \item Performative prediction could arise as a matter of generalization \citep{shah2022goal}, if all training data is historical and the resulting model is then deployed in an environment where it knows it can influence the outcome
    \item Performative prediction could arise if powerful models are developed through a process other than gradient-based updating.
\end{enumerate}
One approach to avoiding performativity is to elicit predictions conditional on different actions that can be taken in response to the prediction. As this reaction is the causal pathway by which a prediction affects outcomes, conditioning on it removes the performative aspect.\\
Unfortunately, eliciting such conditional predictions introduces a new type of performativity. The predictor can influence which action gets taken by dishonestly reporting certain predictions. In fact, it is impossible for decision maker to to deterministically identify and take their most preferred action when eliciting predictions from an agent using a symmetric scoring rule \citep{othman2010decision}.\\
The symmetry of the scoring rule, meaning it is invariant to permutations of outcomes, is important because it avoids the need to establish preferences over all outcomes. This is a famously difficult aspect of the AI alignment problem, due to the difficulty of specifying one's values \citep{hadfieldmenell2018incomplete}. 
\subsection{Our Contribution}
Our contribution is to show that when jointly evaluating two or more predictors, it is possible to elicit honest predictions using a symmetric scoring rule and to use this info to deterministically take the best action. This enables the safe use of conditional predictions from powerful AI systems, either for their own sake or as a strategy to avoid performative prediction.\\
Our primary result is a theorem showing that when predictors engage in a zero-sum competition for accuracy, they have no incentive to influence which action gets taken, and a decision maker can exploit this to elicit honest conditional predictions. This is paired with a uniqueness theorem, showing that only zero-sum competition can make this possible.\\
Supplementary results address obstacles to implementing the mechanism in practice. We show that large spaces of possible actions can be efficiently searched to identify the optimal choice. Additionally, we show that our main results extend to situations where a decision maker chooses stochastically, which adds further incentives for honesty regarding untaken actions.\\
Finally, we show that the mechanism works experimentally in a toy environment. We first show that models trained with a zero-sum objective do not become performative, even in environments incentivizing it. Then we demonstrate zero-sum training untrains performativity from a model faster and to a larger degree than removing the incentives for performativity.
\subsection{Related Work}
The literature on eliciting honest predictions using proper scoring rules is well established (\citealp{Brier1950}; \citealp{Good1952}; \citealp{savage1971elicitation}; \citealp{gneiting2007strictly}). However, it largely dependent on the assumption that making a prediction does not impact the outcome.\\
Concern about predictions affecting outcomes with respect to Oracle AI began in \citep{armstrong2017good}, with a more rigorous formulation and the term ``performative prediction'' being introduced in \citet{perdomo2020performative}. This was followed up with several papers concerning performativity in a single prediction, and largely focused on stability, optimal predictive accuracy, and training methods (\citealp{mendler2020stochastic}, \citealp{izzo2021learn}; \citealp{hardt2022performative}). We focus on performativity across multiple conditional predictions, and address the impact of performative predictions on the welfare of those who elicit them.\\
Using conditional predictions to choose an action was introduced as a decision problem in \citet{othman2010decision}, which went on to prove the impossibility result for using them to take optimal actions. This is the main paper our work responds to, by showing that these results can be overcome using multiple predictors.\\
\citet{chen2011decision} shows that accurate conditional predictions can be elicited if the decision maker is willing to randomize with full support over all actions. However, a human decision maker often cannot commit to randomly taking sufficiently bad actions, and they may not know the exact probabilities they assign to each action as is required. In an ML context, arbitrarily small probabilities require arbitrarily large amounts of training data.\\
\citet{oesterheld2020minimum} shows that a decision maker can take the best action by having the expert choose the action, and reward them proportionally to the resulting utility. This requires specifying their utility function, which is often difficult in general environments. This approach can also fail when outcomes modify the decision maker's utility function or ability to report honestly. Additionally, the information gained from the predicted probabilities can have its own value.
\section{Background and Definitions}
Let $\mathcal{A}$ be a finite set of actions, and let $\mathcal{O}$ be a finite, exhaustive, and mutually exclusive set of outcomes.\\
We start with a decision making principal, who has complete and transitive preferences $\succsim$ over $\Delta(\mathcal{O})$, and $n$ prediction making agents. We say that the principal's preferences uses some tie-breaking procedure, including over actions that produce the same distribution, so their preferences are strict. This simplifies the following theorems and proofs, although it is not crucial for the results.\\
The principal receives a set of predictions $ p \in (\Delta(\mathcal{O})^{|\mathcal{A}| \times n})$, with $p_{i,a,o}$ referring to the probability that agent $i$ assigns to outcome $o$ conditional on action $a$ being chosen. Dropped subscripts indicate all values for the dropped dimension are included, and a negative prefacing a subscript indicates all except that subscript are included. For example e.g. $p_i$ refers to all predictions make by agent $i$ and $p_{-(i,a)}$ refers to all predictions except agent $i$'s predictions for action $a$. After receiving $p$ from the agents, the principal chooses their action using a decision rule $D: (\Delta(\mathcal{O})^{|\mathcal{A}| \times n}) \rightarrow \mathcal{A}$.\\
Once action $a$ is taken and outcome $o$ is realized, agents are assigned scores according to a joint scoring rule $S: \mathcal{A} \times (\Delta(\mathcal{O})^{|\mathcal{A}| \times n})  \times \mathcal{A} \times \mathcal{O} \rightarrow \mathbb{R}^n$, with $S_i$ referring to the function that calculates only agent $i$'s score. 
For notational purposes, we let $S(a, p, o)$ refer to the vector of agent scores after outcome $o$ is realized, and $S(a, p, q)$ refer to the expected scores after the action is taken but before the outcome is realized, where $q \in \Delta(\mathcal{O})^{|\mathcal{A}|}$ represents the true distribution over outcomes. We use $q_a$ to refer to the true distribution conditional on action $a$, and $\textbf{q}$ to refer to the report where all agents predict $q$. For now, we consider the case where the ground truth $q$ is known to all agents. \\
In equilibrium, each agent chooses their report $p_i$ to maximize their expected score $S_i$ conditional on each other agent's report and the decision rule $D$. This means that $p$ is an equilibrium if $\not\exists p_i' \in \Delta(\mathcal{O})$, such that 
\begin{equation*}
    S_i(D((p_i', p_{-i})), (p_i',  p_{-i}), q) \geq S_i(D(p), p, q )
\end{equation*}
In a strong equilibrium, each \textit{group} of agents chooses their reports to maximize their expected scores conditional on the remaining agents' reports and the decision rule $D$. Formally, if $p$ is a strong equilibrium,  $\not\exists C \subseteq \{1,...,n\}, p_C' \in \Delta(\mathcal{O})^{|C|}$, such that $\forall i \in C$,
\begin{equation*}
    S_i(D((p_C', p_{-C})), (p_C',  p_{-C}), q ) \geq S_i(D(p), p, q )
\end{equation*}
with at least one inequality being strict.\\
A joint scoring rule/decision rule pair is \textit{strictly proper} if there exists exactly one equilibrium, and in it each agent reports their true beliefs. Formally, $\forall q, p$, $\exists i, p_i'$ such that
\begin{equation*}
    S_i(D((p_i',  p_{-i})), (p_i',  p_{-i}), q) > S_i(D(p), p, q)
\end{equation*}
if and only if $p \ne \textbf{q}$\\
Let $a^*$ be the principal's most preferred action, so that $q_{a^*} \succ q_{a},\ \forall a \in \mathcal{A} \setminus \{a^*\} $.\\
A joint scoring rule/decision rule pair is \textit{quasi-strictly proper} if there exists at least one equilibrium and in all equilibria $a^*$ is the chosen action, all agent reports their true belief for $a^*$, and all agents are weakly incentivized to report their true beliefs for all actions. Formally, $\forall\ q$, $p = \textbf{q}$ is an equilibrium and $\forall p \ne \textbf{q}$, if either $\exists p_{j, a^*} \ne q_{a^*}$ or $D(p) \ne a^*$, then $\exists i, p_i'$ such that
\begin{equation*}
    S_i(D((p_i',  p_{-i})), (p_i',  p_{-i}), q) > S_i(D(p), p, q)
\end{equation*}
A joint scoring rule/decision rule pair is \textit{symmetric} if (modulo the principal's tie-breaking procedure) they remain consistent under all permutations of the actions, outcomes, and agents.
\section{Theoretical Results}
In the case of a single agent, \cite{othman2010decision} defines the \textit{max} decision rule to be the decision rule that chooses an action $a$ where $p_{1,a} \succsim p_{1, a'}$ $\forall a' \in \mathcal{A}$. A key result of their paper is that there is no symmetric scoring rule that is quasi-strictly proper when combined with the max decision rule. The practical implication of this is that there is no symmetric scoring rule/decision rule pair that deterministically chooses the principal's most preferred action.\\
To provide intuition,\cite{othman2010decision} includes an example with are two actions, $a_1$ and $a_2$, along with two outcomes, $o_1$ and $o_2$. The principal wants to maximize the probability of $o_1$, and we have that $q_{a_1} = [0.5, 0.5]$ and $q_{a_2} = [0.25, 0.75]$. The agent is evaluated with the log scoring rule, which assigns a score of $\log(p_{a,o})$ when action $a$ is chosen and outcome $o$ is realized. \\
If the agent reports honestly for both actions, then action $a_1$ will be chosen by the max decision rule and the agent's expected score will be $0.5 * \log(0.5) + 0.5 * \log(0.5) = \log(0.5)$. If the agent instead reports $p_{1, a_1} = [0.2, 0.8]$ while reporting honestly for $a_2$, then the max decision rule will select $a_2$ and the agent's expected score becomes $0.25\log(0.25) + 0.75\log(0.75) > \log(0.5)$. So, the agent is incentivized to misrepresent the action that the principal would prefer if they knew the truth, causing the principal to choose the lower variance action instead.\\
To begin our own contribution, we start with a couple definitions. A joint scoring rule is \textit{zero-sum} if it has the form 
\begin{equation*}
    S_i(a, p, q) = s(p_{i,a}, q_a) - \frac{\sum_{j \ne i} s(p_{j,a}, q_a)}{n - 1} + c
\end{equation*}
where $c$ is a constant and $s$ is a symmetric, strictly proper scoring rule for the one agent and one action case. For simplicity, we assume $c = 0$ for all following theorems, and we will note when that assumption is relevant.\\
A decision rule is \textit{optimistic} if, for each action, the principal's most preferred prediction conditional on an action is a sufficient statistic for all predictions conditional on that action. Formally, if $p_{-(i,a)} = p_{-(i,a)}'$ and $\exists j$ such that $p_{j,a} \succsim p_{i,a}$ and $p_{j,a} \succ p_{i,a}'$, then $D(p) = D(p')$. An optimistic principal only considers the most preferred prediction they receive for any action.\\
The \textit{optimistic-max} decision rule is the optimistic multi-agent version of the max decision rule, choosing $D(p) = a$ if and only if $\exists i$ such that $p_{i,a} \succsim p_{j, a'},\ \forall j \in \{1,...,n\}, a' \in \mathcal{A}$. As we will see in the following proof, the optimistic-max decision rule is a good analogue for the max-decision rule when paired with a zero-sum scoring rule, as in all equilibria it selects the same action as running the max-decision rule on any individual agent. We restrict our analysis to equilibrium behavior without concern about the ease of computing equilibria, since all agents reporting honestly is always an equilibrium under the mechanisms we analyze.\\
Before we proceed, we first introduce a useful lemma.
\begin{lemma}\label{lemma1}
    Under a zero-sum scoring rule $S$, all agents receive an expected score of $0$ in any equilibrium.
\end{lemma}
The proofs for this and all theoretical results are provided in the technical appendix.\\
Using this lemma, we can construct a decision rule/scoring rule pair that allows for the the Othman and Sandholm (2010) impossibility result to be overcome.
\begin{theorem}
When $n \geq 2$, the combination of the optimistic-max decision rule $D$ and a zero-sum scoring rule $S$ is quasi-strictly proper, and in any equilibrium the max decision rule applied to any agent selects action $a^*$.
\end{theorem}
As this is our key contribution, and the proof provides substantial intuition as to why the result holds, we provide it below.
\begin{proof}
First, we show that in equilibrium, $\not \exists p_{i, a}$ such that $p_{i, a} \succ q_{a^*} $. Suppose $p$ is an equilibrium, and such a prediction exists. Based on the decision rule, the principal must end up choosing some action $a'$ where $\exists p_{j, a'} \succ q_{a^*}$. Then, since the decision rule is optimistic, there exists some agent $k \ne j$ who is either reporting honestly or can change their prediction to $p_{k,a'} = q_{a'}$ without affecting the action taken. The score for such a prediction, $S_k(a', (q_{a'}, p_{-(k,a')}), q )$, is equal to
\begin{equation*}
    s(q_{a'}, q_{a'}) - \frac{ s(p_{j, a'}, q_{a'})}{n - 1} - \frac{\sum_{i \ne j,k} S'(p_{i, a'}, q_{a'})}{n - 1} > 0
\end{equation*}
The inequality follows because $s(\dot, q_{a'})$ is uniquely maximized at $q_{a'}$, and $p_{j, a'} \ne q_{a'}$. By Lemma \ref{lemma1}, this contradicts that $p$ is an equilibrium.\\
Next, we show that in equilibrium, $\not \exists i$ such that $q_{a^*} \succ p_{i, a^*}$. Suppose $p$ is an equilibrium, and such a prediction exists. If another agent $j \ne i$ reports honestly, then $D(p) = a^*$ since the decision rule is optimistic and we have previously established that no predictions are more preferred to $q_{a^*}$. The score for such a prediction, $S_k(a', (q_{a^*}, p_{-(j,a^*)}), q )$, is equal to
\begin{equation*}
    s(q_{a^*},q_{a^*}) - \frac{ s(p_{i, a'}, q_{a^*})}{n - 1} - \frac{\sum_{k \ne i,j} S'(p_{k, a^*}, q_{a^*})}{n - 1} > 0
\end{equation*}
By Lemma \ref{lemma1}, this contradicts that $p$ is an equilibrium.\\
In equilibrium, each agent reports honestly for $a^*$ and there are no reports $p_{i,a} \succ q_{a^*}$, so running the max decision rule on any $p_i$ must choose $a^*$. Using the optimistic-max decision rule across predictors similarly chooses $a^*$. Predictions conditional on untaken actions do not enter the scoring function, and so honesty is weakly incentivized. As such, the decision/scoring rule pair is quasi-strictly proper.
\end{proof}
The intuition for why this works is that from each agent's perspective, the other predictions are constant. Therefore, they face a strictly proper scoring rule and are incentivized to predict honestly. Affecting the principal's choice of action and the resulting distribution of outcomes provides no benefit to them. Since scores are zero-sum, any score increase from a distribution shift would be exactly offset by an equivalent score increase to the other agents. The optimistic decision rule ensures that if agents are overstating a suboptimal action's value, then at least one agent can report honestly without changing the chosen action, and that if agents are understating the optimal action's value, then at least none agents can report honestly and ensure it is chosen.\\
Optimism is not a necessary property for the decision rule. As we will cover in the subsection on stochastic choice, randomly choosing between which agents to believe also leads to equilibria where the principal deterministically chooses their most preferred action.\\
However, both optimism and stochastic choice can be difficult for the principal to commit to in the event of off-equilibrium behavior. If agents provide conflicting predictions, the principal may not be willing to throw away half of the information provided. Fortunately, if we restrict to principals with preferences that follow the Independence axiom, then we can use a decision rule that incorporates all predictions. The \textit{Independence} axiom states that for any $a,b,c \in [0,1]^{|\mathcal{O}|}, p \in (0,1]$, $a \succ b$ iff $pa + (1-p)c \succ pb + (1-p)c$. Preferences that meet this criteria include those that can be represented by a von Neuman-Morgenstern utility function, as well as lexicographic preferences. The \textit{mean-max} decision rule applies the max decision rule to the mean predictions for each action.
\begin{theorem}
    When $n = 2$, for a principal with preferences that follow Independence, the combination of the mean-max decision rule and a zero-sum scoring rule is quasi-strictly proper
\end{theorem}
The reason for the $n = 2$ restriction is that the proof relies on any single agent being able to counteract a negative misrepresentation of $a^*$ from other agents to ensure that the action is chosen regardless. With $n \geq 3$ agent, this is not always possible. There can be an equilibrium where $q_{a^*} \succ p_{i, a^*}$, $\forall i$, but $\not \exists \bar{p}, \underline{p}$ such that agent $j$ predicting $p_{j, a^*} = \bar{p}$ and $p_{j, a} = \underline{p}$ $\forall a \ne a^*$ results in $a^*$ being chosen. However, if the agents can collaborate, or if the principal can make stochastic choices (see Theorem \ref{random-mean-max}), the result also holds for larger groups of agents.
\begin{theorem}
    When $n \geq 2$, for a principal with preferences that follow Independence, the combination of the mean-max decision rule and a zero-sum scoring rule is quasi-strictly proper when restricting to strong equilibria.
\end{theorem}
The idea here is that even if any individual agent cannot convince the principal to choose a certain action, all but one working together can. So, if any agent is dishonest in a way that would result in $a^*$ not being chosen, all the other agents can exploit that to earn a positive score.\\
A nice property of these quasi-strictly proper mechanisms is that beyond having no incentive to lie about untaken actions, there \textit{is} an incentive not to lie too drastically. If there exists a prediction $p_{j, a'}$ for untaken action $a'$ for which $s(p_{j, a'}, q_{a'})$ is sufficiently low, then for $i \ne j$ there exists a prediction $p_{i, a'} \succ q_{a^*}$ that results in action $a'$ being chosen and agent $i$ receiving a positive expected score, along with agent $j$ receiving a negative one.\\
The choice of scoring rule affects how inaccurate a prediction for an untaken action can be in equilibrium. For example, $p_{i,a',o} = 0$ when $q_{a,o} > 0$ results in an expected score of negative infinity under the log scoring rule, and only a finitely bad score under the Brier scoring rule, so whether that prediction is possible in equilibrium depends on the scoring rule chosen. The optimal choice to minimize inaccuracy depends on the principal's preferences and beliefs about the distribution of $q$.\\
Beyond merely limiting the size of divergence from honest reporting for unchosen actions, it is possible to strictly incentivize honesty for all actions. Consider a \textit{disagreement-seeking-max} decision rule, which always chooses an action for which predictions differ if possible, and follows the max decision rule otherwise. Formally,  if $\tilde{A}_{p} \equiv \{a \in \mathcal{A}| \exists i,j\ s.t.\ p_{i,a} \ne p_{j,a} \} \ne \emptyset$ then $D(p) \in \tilde{A}_p$, and otherwise $D(p) = \hat{a}$, where $p_{i,\hat{a}} \succ p_{i,a'}$, $\forall a' \in \mathcal{A} \setminus \{\hat{a}\}$.
\begin{theorem}
    If $n \geq 2$, the combination of a disagreement-seeking-max decision rule and a zero-sum scoring rule is strictly proper, and the principal deterministically chooses $a^*$.
\end{theorem}
The downside of using a disagreement-seeking-max decision rule is that the proof for the above theorem is heavily reliant on the assumption that all agents know the ground truth $q$. If any of them make even slight errors, or if there is a small amount of noise in their report, then the rule instead selects a random action, with no relation to the principal's preferences. This contrasts the optimistic-max and mean-max decision rules where adding a small amount of noise to either agent's prediction will at worst still result in choosing a close to optimal action.\\
There are multiple decision rules that can result in a quasi-strictly proper mechanism when paired with a zero-sum scoring rule. How important is the zero-sum property in the scoring rule? It can show that zero-sum scoring rules are unique in being able to achieve the incentive for honesty.
\begin{theorem}
    If a symmetric scoring rule/decision rule pair is quasi-strictly proper, then the scoring rule is zero-sum.
\end{theorem}
This is an important result, because it drastically reduces the amount of space through which to search for the optimal scoring rule/decision rule pair. It is also highly relevant to AI alignment, as it means that incentives for honesty are highly unlikely to arise accidentally or by default. The explicit implementation of a zero-sum scoring rule is necessary to achieve that end.
\subsection{Efficient Search}
If a decision problem is being used to avoid performative prediction, one danger is that the action space is not sufficiently descriptive. For example, a professor going for lunch could divide the action space into the nearby restaurants without clarifying what they would order at each one. That reintroduces the possibility of performative prediction by affecting the choice of meal, for example by influencing them to choose a cheap option by predicting they will not enjoy it. While any division of the action space reduces a prediction's influence on the outcome, ideally we would like that influence to be as small as possible. To that end, we could end up with an enormous action space, and very small details distinguishing different actions. Fortunately, even a large action space can be searched efficiently.
\begin{theorem}
    A principal can identify $a^*$ with at most $O(log(|\mathcal{A}|))$ comparisons between actions.
\end{theorem}
\begin{algorithm}[t!]
\caption{This algorithm performs a binary search over actions, using conditional predictions to narrow the action space}
\label{alg:binary-action-search}
\begin{algorithmic}[1]
\Procedure{BinaryActionSearch}{$\mathcal{A}, D, S$}
    \While{$|\mathcal{A}| > 1$}
        \State $A_1 \gets \text{half of } \mathcal{A}$
        \State $A_2 \gets \text{other half of} \mathcal{A}$
        \State $p_{A_1}, p_{A_2}  \gets \text{ElicitPrediction}(A_1, A_2)$
        \State $\mathcal{A} \gets D((p_{A_2}, p_{A_2}))$
    \EndWhile
    \State \Return $\text{single action in } A$
\EndProcedure
\Procedure{ElicitPrediction}{$A_i, A_j$}
    \State \Return predictions from $n$ agents evaluated with $S$ if $A_i$ or $A_j$ is selected
\EndProcedure
\end{algorithmic}
\end{algorithm}
This proof shows that Algorithm \ref{alg:binary-action-search}, which essentially does a binary search through action space, maintains the incentives for honesty even as predictions can affect later predictions. Here, we rely on setting the constant term $c$ of the zero-sum scoring rule to zero.\\
One way of implementing Algorithm \ref{alg:binary-action-search} would be to add more and more details to the action with each iteration. For example, when deciding on lunch plans, one could first ask what cuisine they would most enjoy, then what restaurant, then which menu item, accumulating additional information with each step. Iteratively eliciting information is not necessary, however, and it is possible to jump straight to the end of the process. 
\begin{theorem}
    A principal can identify $a^*$ with at most $O(1)$ comparisons between actions.
\end{theorem}
The way this works is by eliciting a non-conditional, non-zero-sum prediction from an agent about which action will be chosen, then running Algorithm \ref{alg:binary-action-search} but with the action space instead split into the set containing only that action and the set containing all other actions. $a^*$ is then identified after a single comparison.
\begin{figure*}
  \centering
  \includegraphics[width=\textwidth]{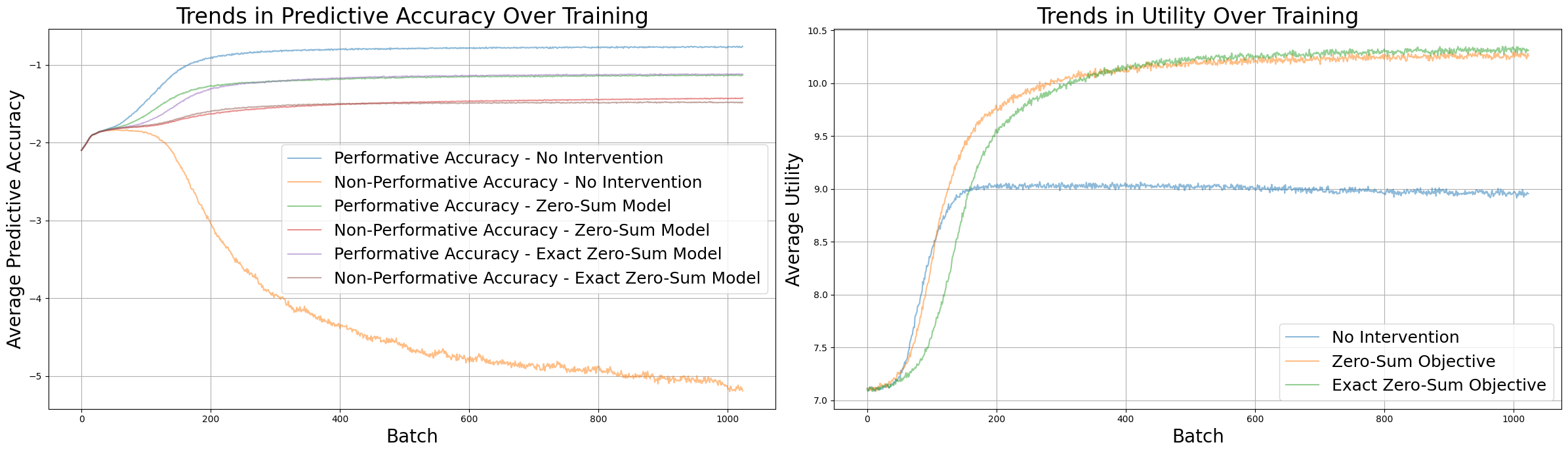}
  \caption{Figure 1: (Left) In environments that incentivize performative prediction, training with a zero sum objective avoids the model becoming performative, and results increasing accuracy across predictions. Models trained with no intervention are more accurate for whichever action is chosen, as they influence the choice to be easier to predict, but are less accurate overall. (Right) When no intervention prevents a model from becoming performative, user utility plateaus earlier and at a lower level.}
  \label{fig:experiment1}
\end{figure*}
\subsection{Stochastic Choice}
The principal may be interested not only in identifying $a^*$, but in the information contained in predictions about untaken actions. For example, consider training an AI system that generates potential actions in the way LLMs generate potential text completions. We would like to have a reward model evaluate the outcomes of each action, but in an online training environment only one can be taken, so we instead evaluate predictions. Performative prediction would be disastrous, as it would train the model to generate actions that are easy to predict rather than actions that the developers desire, but even a quasi-strictly proper scoring rule is insufficent. We would also want accurate predictions for the untaken actions, so they can be properly evaluated.\\
If the principle willing to partially randomize their decision, then for reasonable methods of randomization, the actions for which honest predictions are strictly incentivized can be greatly expanded. This can be done with arbitrarily small probability of not taking $a^*$, and with zero chance of taking deeply unpalatable actions.\\
Let $D: (\Delta(\mathcal{O})^{|\mathcal{A}|})^n \rightarrow \Delta(\mathcal{A})$ be a non-deterministic decision function, with $D_{a}$ representing the probability assigned to action $a$. Consider the following regularity conditions on $D$:
\begin{condition}\label{condition_1}
    If $p_{i,a}' \succ p_{i,a}'$ $\forall a \in A \subseteq \mathcal{A}$ and $p_{-(i,A)} = p_{-(i,A)}'$, then $\exists a \in A$ such that $D_{a}(p) > 0$, implies $\exists a' \in A$ such that $D_{a'}(p') > 0$
\end{condition}
\begin{condition}\label{condition_2}
    If $p_{-(i,a)} = p_{-(i,a)}'$, $p_{i,a} \succ p_{i,a}'$ then for $a' \ne a$ $D_{a'}(p) > 0$ implies $D_{a'}(p') > 0$
\end{condition}
Condition \ref{condition_1} says that if an agent's predictions for some subset of actions are all changed to more preferred distributions, then if at least one action in that subset was assigned positive probability before the change, at least one will be assigned positive probability afterwards. Condition \ref{condition_2} says that if an agent's prediction for some action changes to a less preferred distribution, this alone will not cause the principal to assign zero probability to a different action. \\
These conditions ensure that in equilibrium, $S_i(a, p, q) = 0$, $\forall a$ s.t. $D_a(p) > 0$. This is distinct from Lemma \ref{lemma1}, which only showed that that $E_{a \sim D(p)}[S_i(a,p,q)] = 0$ in equilibrium.
\begin{lemma}
    Under a zero-sum scoring rule $S$ and optimistic decision rule $D$, if Conditions \ref{condition_1} and \ref{condition_2} are met then in equilibrium $p$ $\forall i, \forall a$ such that $D_a(p) > 0$, $p_{i,a} = q_a$
\end{lemma}
This suggests another decision rule, besides the optimistic-max and mean-max, that can deterministically take $a^*$. We call the decision rule that randomizes which agent to believe and then takes their most preferred prediction the \textit{random-max} decision rule.
\begin{theorem}
    When $n \geq 2$, a zero-sum scoring rule and the random-max decision rule is quasi-strictly proper.
\end{theorem}
We can also incorporate more information from the predictions with a variant on the mean-max decision rule. A \textit{random-mean-max} rule decides according to the mean-max rule with probability $1 - \epsilon$, and excludes each agent from the mean with probability $\frac{\epsilon}{n}$.
\begin{theorem}\label{random-mean-max}
    When $n \geq 2$, a zero-sum scoring rule and the random-mean-max decision rule is quasi-strictly proper.
\end{theorem}
In practice, we may be willing to accept some chance of taking non-optimal actions in order to gather information from the predictions for these actions. While \citet{chen2011decision} showed that randomizing over actions with full support can incentivize predictions for all actions, we show that with multiple agents we can randomize with only partial support and still strictly incentivize honest predictions for those actions. In particular, the principal can restrict support to only the actions they would be willing to randomize over if they knew the true distribution. This means they do not need to commit to taking to catastrophically bad actions with any probability.\\
To do so, we require one more condition:
\begin{condition}
    If $p_{-(i,a)} = p_{-(i,a)}'$ and $D_{a}(p) = D_{a}(p') = 0$ then $D(p) = D(p')$
\end{condition}
This says that if an agent modifying their prediction for an action does not changing the fact that it is assigned zero probability, the probabilities assigned to other actions do not change.
\begin{theorem}
    Under a zero-sum scoring rule $S$ and optimistic decision rule $D$ meeting Conditions 1-3, then in any equilibrium $p$, $D(p) = D(\textbf{q})$
\end{theorem}
This result is particularly useful if we want to train a model to take actions that are predicted to have good outcomes. It allows us to strictly incentivize honesty for relevant untaken actions, while ensuring safe exploration. Knowing that some actions are too bad to be considered is also useful knowledge, even without being guaranteed honest predictions for them.
\begin{figure*}
  \centering
  \includegraphics[width=\textwidth]{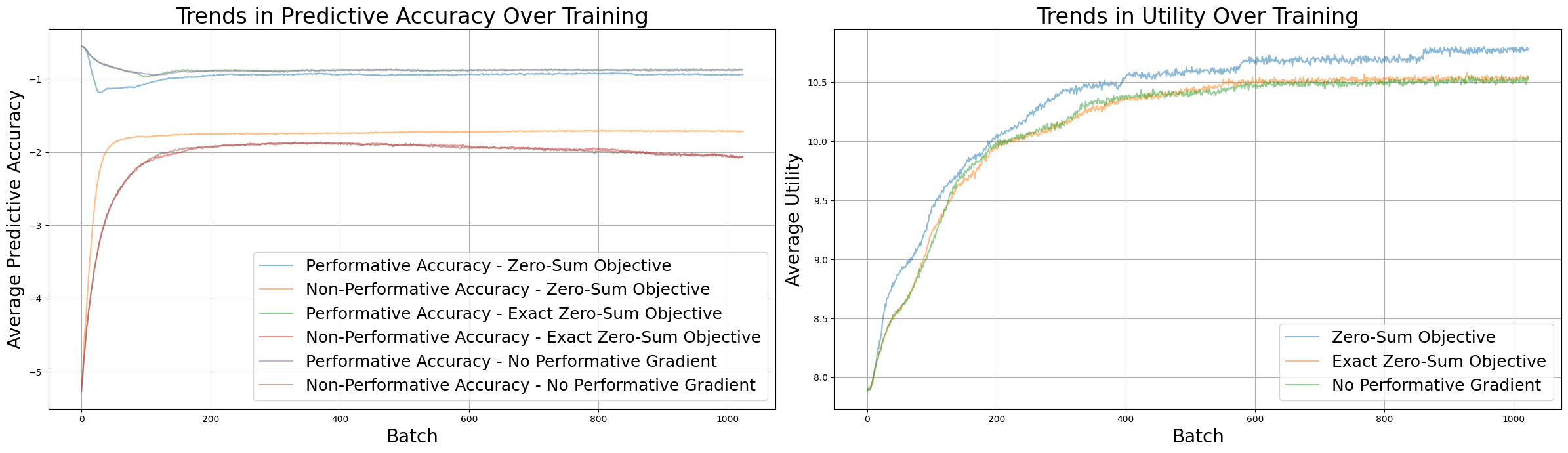}
  \caption{Figure 2: (Left) A zero-sum objective using two different predictions from the same model results in faster and larger decreases in performativity than an exact zero-sum objective or training in a non-performative environment. (Right) The decrease in performativity leads to higher utility for users, with larger gains for larger drops in performativity.}
  \label{fig:experiment2}
\end{figure*}
\section{Experiments}
We test our theoretical results in a toy environment, with eight possible actions, eight possible outcomes, and eight variables representing context. The ground truth probabilities are given by a randomly initialized neural net that takes in as input the context and a one-hot vector representing the choice of action, and outputs a distribution over outcomes. A principal with a randomly generated utility function over actions makes their decisions by taking the softmax of their expected utility from each action, consolidating reports from agents optimistically.\\
We train models to predict the outcome, using a cross-entropy loss function and running the gradient through the impact of the prediction on the principal's decision. This is the simplest way to implement performativity in a toy environment, and shows the robustness of zero-sum competition in avoiding it.\\
Our first experiment compares training with no intervention to two methods of implementing a zero-sum objective. The first method trains an agent against a detached version of itself that makes identical predictions, which we call \textit{exact}. The second uses dropout to generate two different predictions from the same model, then performs a gradient update for each one while detaching the other. This provides evidence on behavior when agents have different information and thus disagree about probabilities. We apply dropout to all models in order to isolate the effect of a zero-sum objective.\\
In Figure \ref{fig:experiment1} we compare how the various training methods affect predictive accuracy, both when taking performativity into account and when weighing the predictions for all actions equally. We also measure how the principal's utility changes throughout the training process.\\
We can see that both implementations of a zero-sum objective perform very similarly, increasing in both measures of predictive accuracy. Performative predictive accuracy is slightly higher, as higher utility actions tend to have more extreme distributions, which results in a higher prediction score.\\
In contrast, training without an intervention leads to the largest gain in performative predictive accuracy, but after an initial increase non-performative predictive accuracy drops off sharply. When this divergence occurs, the principal's utility plateaus, whereas for the zero-sum objectives it continues to rise.\\
Without an intervention, performativity compounds throughout training. The more inaccurate the conditional prediction for an action is, the more the local gradient pushes towards performativity to ensure that action is not taken. Similarly, it discourages gradient updates from making the prediction more accurate if doing so increases the chance the action will be chosen.\\
Our second experiment tests whether a model that has already become performative can have that behavior trained out of it. We compare the same implementations of a zero-sum objective, alongside removing the gradient that runs through the principal's choice of action. This last intervention can be thought of as training to predict only historical data, rather than making predictions that can affect their own outcome.\\
Figure \ref{fig:experiment2} compares the different training methods using the same measures as in the first experiment. We can see that the exact zero-sum objective behaves like training in a non-performative environment, which makes sense since they produce nearly identical gradients. The zero-sum objective that generates two distinct predictions untrains performativity faster, plateaus at a higher level of predictive accuracy, and results in higher utility for the principal. We speculate that this results from a wider range of actions being selected throughout training. \\
We further run robustness checks to ensure that the results are not affected by experimental choices. No major changes were observed after changing the decision rule from optimistic to mean, only assigning positive probability to above-median expected utility actions, changing the scoring rule base from log score to Brier score, sampling more than two agents when calculating the zero-sum objective, or pretraining the model on historical data. These results are available in the technical appendix.
\section{Discussion and Future Direction}
Our work demonstrates that it is possible to incentivize honesty in conditional predictions, both theoretically and practically. This allows for the elicitation of conditional predictions as a safe alternative to accepting performativity in the unconditional case.\\
Eliminating the risk of performative prediction from powerful AI systems represents a major step towards their safe usage. This applies both to Oracle AI as a strategy for aligning superintelligent systems, and to less advanced systems which may be used either for economic purposes or in the training of more powerful systems. Honesty means we can trust the predictions they output, putting the onus on us to use that information wisely.\\
There are two main directions we are interested in expansions of this work. The first is empirical, as we have only demonstrated the success of zero-sum training in incentivizing honesty in a toy environment. Extending these results to more complex environments, especially one using real world data, would give us additional confidence in the robustness of the result.\\
The second direction is theoretical, loosening the assumption that all agents have the same information. While this is less concerning when both agents are contained in the same model, and our experiments showed this does not result in dishonesty, it would provide theoretical reassurance that the incentives are always aligned for any two agents.\\
Once we can conclusively eliminate concerns about performative prediction, we can move on to the next question: how can we use honest predictions to align powerful AI systems and ensure a safer world?

\nocite{*}

\end{document}